# Rethinking Scanning Strategies with Vision Mamba in Semantic Segmentation of Remote Sensing Imagery: An Experimental Study

Qinfeng Zhu, *Graduate Student Member, IEEE,* Yuan Fang, *Graduate Student Member,* Yuanzhi Cai, *Member, IEEE,* Cheng Chen, Lei Fan, *Member, IEEE*

*Abstract*—Deep learning methods, especially Convolutional Neural Networks (CNN) and Vision Transformer (ViT), are frequently employed to perform semantic segmentation of high-resolution remotely sensed images. However, CNNs are constrained by their restricted receptive fields, while ViTs face challenges due to their quadratic complexity. Recently, the Mamba model, featuring linear complexity and a global receptive field, has gained extensive attention for vision tasks. In such tasks, images need to be serialized to form sequences compatible with the Mamba model. Numerous research efforts have explored scanning strategies to serialize images, aiming to enhance the Mamba model's understanding of images. However, the effectiveness of these scanning strategies remains uncertain. In this research, we conduct a comprehensive experimental investigation on the impact of mainstream scanning directions and their combinations on semantic segmentation of remotely sensed images. Through extensive experiments on the LoveDA, ISPRS Potsdam, and ISPRS Vaihingen datasets, we demonstrate that no single scanning strategy outperforms others, regardless of their complexity or the number of scanning directions involved. A simple, single scanning direction is deemed sufficient for semantic segmentation of high-resolution remotely sensed images. Relevant directions for future research are also recommended.

## I. INTRODUCTION

S<small>EMANTIC</small> segmentation, a typical task for remotely sensed imagery, involves classifying every pixel in an image into categories such as roads, vegetation, buildings, and water bodies [1]. In recent years, deep learning techniques have significantly advanced the accuracy of semantic segmentation of remotely sensed images, especially when handling complex patterns and diverse variations inherent in remote sensing scenes [2]. Specifically, Convolutional Neural Networks (CNNs) [3] and Vision Transformers (ViTs) [4] are now commonly used backbone networks for semantic segmentation of remotely sensed images, evidenced by continuous evolution of methods that deliver state-of-the-art performance [5-8].

However, these widely employed network architectures also face challenges when segmenting high-resolution remotely sensed images. CNNs, due to their limited receptive fields, struggle to capture long-range semantic dependencies present within high-resolution images [9]. Although ViTs possess a global receptive field, their quadratic complexity makes them challenging to deploy for high-resolution images [10]. To address these challenges, researchers have started to investigate architectures based on a newly introduced network called Mamba [11].

Mamba, a network based on State Space Models (SSMs) [12], was initially applied to large language models [13, 14]. Mamba functions as a sequential network similar to a Recurrent Neural Network (RNN), capable of inducting prior information and predicting subsequent states. It efficiently compresses long-term contextual information by incorporating a selective mechanism that selectively attends to or ignores inputs. When applied

This work was supported in part by the Xi'an Jiaotong-Liverpool University Research Enhancement Fund under Grant REF-21-01-003, and in part by the Xi'an Jiaotong-Liverpool University Postgraduate Research Scholarship under Grant FOS2210JJ03.
Corresponding author: Lei Fan

Qinfeng Zhu, Yuan Fang, Cheng Chen, and Lei Fan are with the Department of Civil Engineering, Xi'an Jiaotong-Liverpool University, Suzhou, 215123, China. (e-mail: Qinfeng.Zhu21@student.xjtlu.edu.cn; Yuan.Fang16@student.xjtlu.edu.cn; Cheng.Chen19@student.xjtlu.edu.cn; lei.fan@xjtlu.edu.cn)
Yuanzhi Cai is with the CSIRO Mineral Resources, Kensington, WA 6151, Australia. (e-mail: Yuanzhi.Cai@CSIRO.AU)

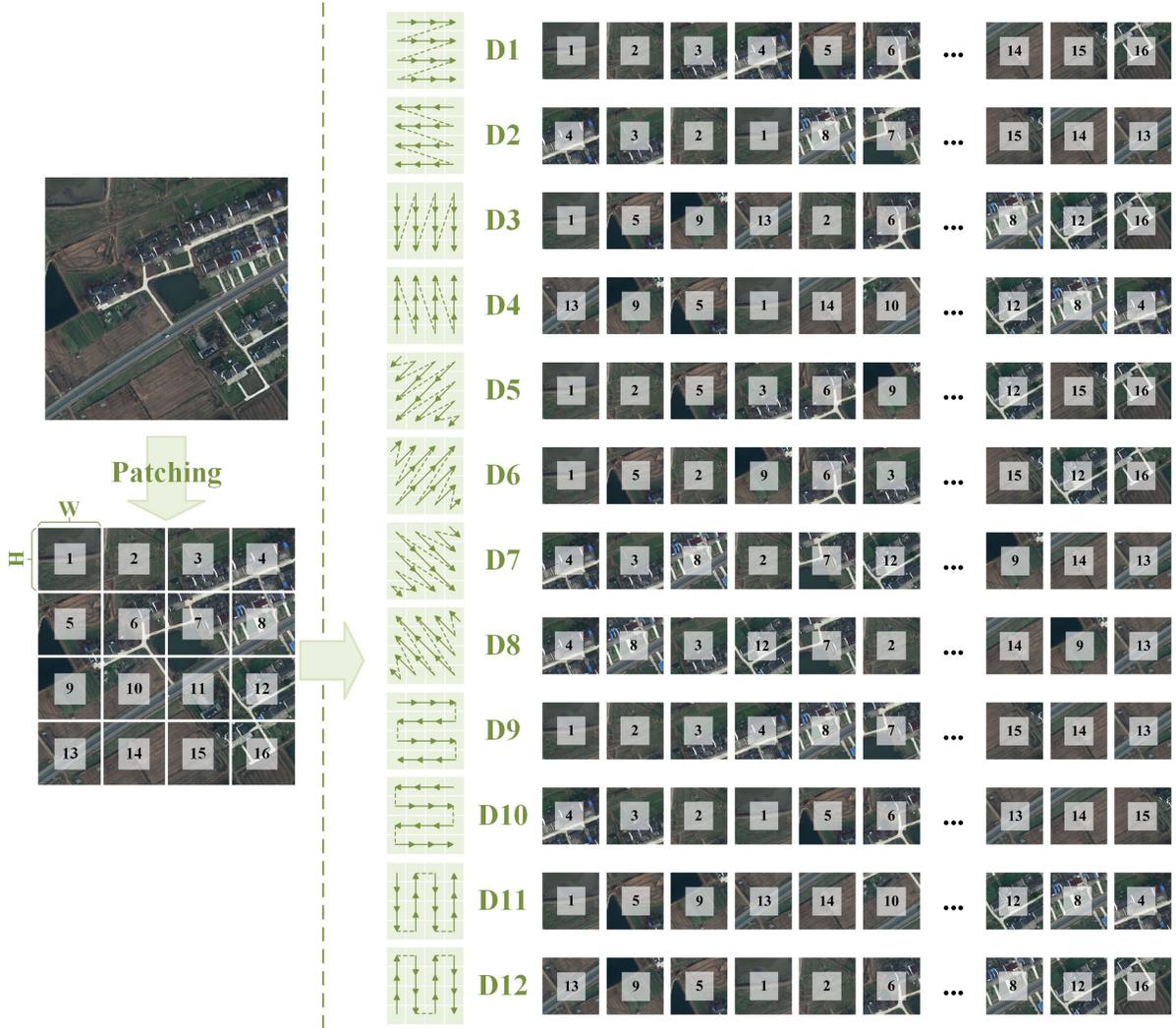

**Fig. 1.** 12 commonly used scanning directions in Vision Mamba. Images are cropped into patches according to a predefined size, and these patches are then modeled into sequences based on specific scanning direction(s).

to vision tasks, this network can achieve a balance between a global receptive field and linear complexity [15], indicating great promise in the segmentation of remotely sensed images.

Drawing on the success of ViTs [4], which introduced the transformer architecture to vision tasks, extensive research [16-18] has successfully integrated mamba into image processing tasks. Similar to ViT, which crops an image into patches and flattens them to be fed into the transformer model, Mamba process flattened image patches as sequences. However, unlike ViT, which computes multi-head self-attention among these image patches, Mamba processes image patches sequentially. Therefore, numerous scanning directions [16-20] of image patches are available.

Extensive research has explored new scanning directions and their combinations, attempting to enhance Mamba's performance in understanding of images. Fig. 1 displays 12 commonly used scanning directions (D1 - D12). D1 - D4 involves sequential scanning every row or column of image patches in a "Z"-shaped pattern. D5 - D8 involve sequential scanning of image patches in diagonal directions. D9 - D12 perform "S"-shaped serpentine scanning of image patches. However, existing studies have not comprehensively compared their effectiveness. Therefore, there is an urgent need for a comparative study to quantitatively evaluate the impact of various scanning directions and their combinations on the performance of Mamba-based methods in typical remote sensing tasks such as semantic segmentation.

In this paper, we designed an experimental framework aimed at undertaking a comprehensive and fair comparison of various scanning strategies, tailored specifically for the semantic segmentation of high-resolution remotely sensed images. We perform 22 scanning strategies in our comparative experiments, including 12 individual scanning directions and 10 combinations of scanning directions. Each scanning strategy is tested across the LoveDA [21], ISPRS Potsdam, and ISPRS Vaihingen datasets.

The main contributions of this paper are as follows.

1. This paper summarizes commonly used scanning strategies of Vision Mamba.
2. For the first time, this study quantitatively assesses the influence of various scanning strategies of Vision Mamba on the accuracy of semantic segmentation of remotely sensed imagery, using a specifically-designed experimental framework.

## II. RELATED WORK

*A. Development of State Space Model*

Recurrent Neural Networks (RNNs) [22] are classic sequential networks that calculate the current hidden state $h_t$ based on the previous hidden state $h_{t-1}$ and the current input $x_t$, thus predicting the output $y_t$ of the current state. However, as the sequence progresses, RNNs tend to forget earlier information and cannot be trained in parallel. In contrast, SSMs can predict the current output based on a summary of all previous states and the current input, avoiding the problem of forgetting. Specifically, SSMs can be expressed using Eqn.(1) and Eqn.(2).

$$h'(t) = Ah(t) + Bx(t) \tag{1}$$

$$y(t) = Ch(t) + Dx(t) \tag{2}$$

where $x(t)$ represents the input at the current state. $h(t)$ represents the previous hidden state. $h'(t)$ represents the update to the hidden state. $y(t)$ represents the prediction for the current state.

However, such SSMs are unable to handle discrete data inputs. Since many data types are discrete, enabling SSMs to cope with discrete data is meaningful. This can effectively be achieved by the Structured State Space for Sequences (S4) [23]. It employs the zero-order hold technique to discretize the SSM, which can be described through Eqn.(3) - Eqn.(7).

$$h_k = \bar{A}h_{k-1} + \bar{B}x_k \tag{3}$$

$$y_k = \bar{C}h_k + \bar{D}x_k \tag{4}$$

$$\bar{A} = e^{\Delta A} \tag{5}$$

$$\bar{B} = (e^{\Delta A} - I)A^{-1}B \tag{6}$$

$$\bar{C} = C \tag{7}$$

where $\bar{A}$ and $\bar{B}$ represent the discretized matrices $A$ and $B$, respectively. $h_{k-1}$ denotes the previous hidden state, and $h_k$ represents the current hidden state. Additionally, S4 incorporates a High-order Polynomial Projection Operator [24] to address the issue of long-distance dependencies in sequence modeling.

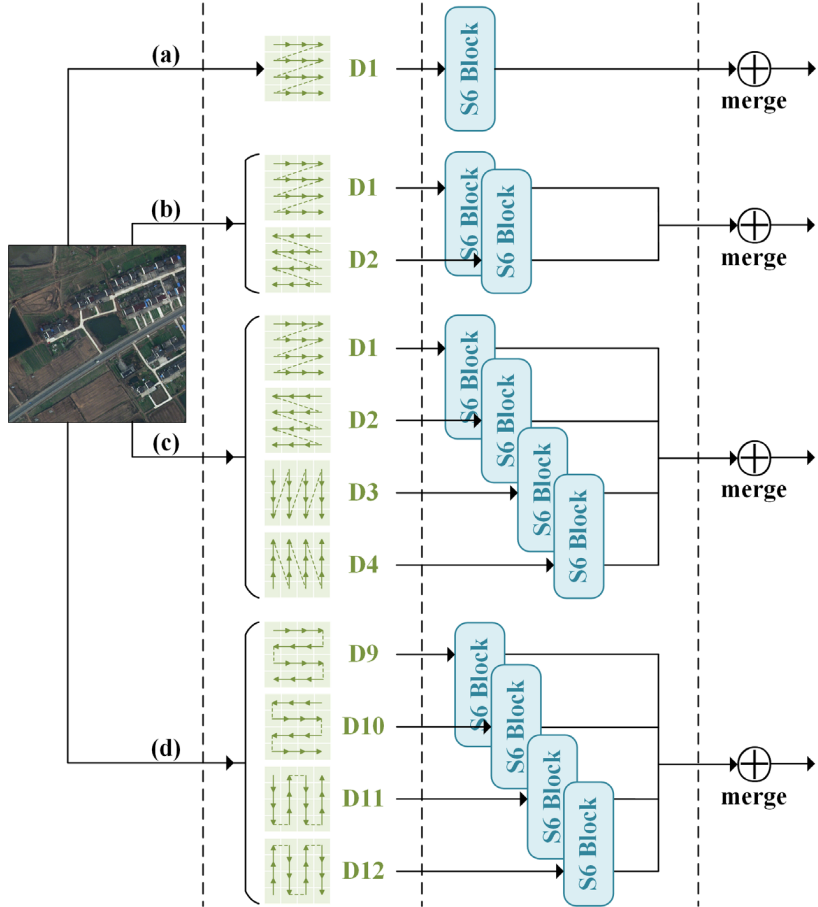

**Fig. 2.** (a) Flattening scanning strategy, consistent with Samba; (b) Scanning forward and backward after flattening, followed by merging, consistent with Vim; (c) Scanning in four directions sequentially, followed by merging, consistent with VMamba; (d) Serpentine scanning in four directions, consistent with PlainMamba.

However, due to the issue of linear time invariance [25], the matrices $A$, $B$, and $C$ in S4 cannot adaptively infer based on different inputs. Specifically, in natural language processing, S4 treats each word in a sentence with the same level of attention, unable to focus selectively on more important parts. Integrating a selection mechanism into S4 enables selective information processing in the Mamba architecture, allowing it to prioritize or disregard specific inputs. This capability facilitates an effective summary of previous information. Extensive experimental studies have demonstrated the advantages of the Mamba architecture in natural language processing [11].

*B. Development of Vision Mamba*

Since Mamba is a sequential network that cannot directly process two-dimensional image data, exploring methods to serialize images is meaningful. The first attempt Vim [17], which is similar to ViT [4], involves cropping an image into patches and flattening them. It performs both forward (D1) and reverse (D2) scans of image patches in rows before merging them, as shown in Fig. 2(b). Similarly, VMamba [16] builds on the foundation of ViM by adding two vertical scanning directions (D3, D4), as shown in Fig. 2(c). PlainMamba [18] adopts a serpentine scanning approach (D8, D9, D10, D11), as illustrated in Fig. 3(d).

These efforts are all based on the hypothesis that varying scanning directions of image patches can potentially enhance Mamba's understanding of images. However, there is a lack of comprehensive and quantitative comparisons of model performances under different scanning directions in their work. For example, Vim and PlainMamba lacked essential ablation studies to validate their scanning methods. In VMamba, the results for horizontal scan combinations (i.e., D1 and D2) were not reported, while the four-directional scanning (i.e., D1, D2, D3 and D4) achieved only a 0.3% higher accuracy on ImageNet [3] compared to the unidirectional scanning D1. Considering the likely fluctuations in model performance during training, this marginal improvement is inadequate to confirm the effectiveness of multidirectional scanning.

*C. Mamba-based Semantic Segmentation*

As Vision Mamba continues to evolve, numerous studies have been undertaken to assess its performance in semantic segmentation tasks, particularly in the domain of medical imaging and remote sensing. In these studies, different scanning strategies were also considered to test their impact on Mamba's ability of image understanding.

U-Mamba [26] represented the first attempt to merge Mamba with the UNet [27] architecture for semantic segmentation of medical images. However, due to its simplistic architectural design, its performance fell short of the then state-of-the-art segmentation methods. Subsequently, several enhanced methods [28-32] emerged, using bi-directional scanning with Vim and/or four-directional scanning with VMamba.

In the context remote sensing, Samba [20] was the first study that introduces Mamba into semantic segmentation of remotely sensed images, in which image patches are flattened in the same manner as ViT, as shown in Fig. 2(a). Later, RS3Mamba [33] used a four-directional scanning method of VMamba to construct an auxiliary encoder for semantic segmentation. Similarly, RSMamba [19] expanded on VMamba's four-directional scanning by adding four additional diagonal directions (i.e., D5, D6, D7 and D8) in its encoder-decoder architecture.

### III. EXPERIMENTAL FRAMEWORK

To thoroughly assess the impact of scanning strategies on Mamba's performance in semantic segmentation tasks with high-resolution images, we have designed a specific semantic segmentation framework using an encoder-decoder architecture to facilitate quantitative comparisons of scanning strategies.

*A. Overall Architecture*

The overall framework is shown on the left-hand side of Fig. 3. Images are divided into patches in the encoder section, and then sequentially fed into four Vision Mamba Scan (VMS) blocks for progressive downsampling. To ensure the fairness of the experiments, we consistently use UperNet [34], the state-of-the-art network for segmentation, as the decoder for producing segmentation results.

*B. VMS Block*

The VMS Block is a residual network with skip connections. The residual network consists of two branches. One branch uses a depth-wise convolution (DW Conv) layer to extract features, performs S6 calculations [11] on scans in various directions, and subsequently merges them. The other branch consists of a linear mapping followed by an activation layer.

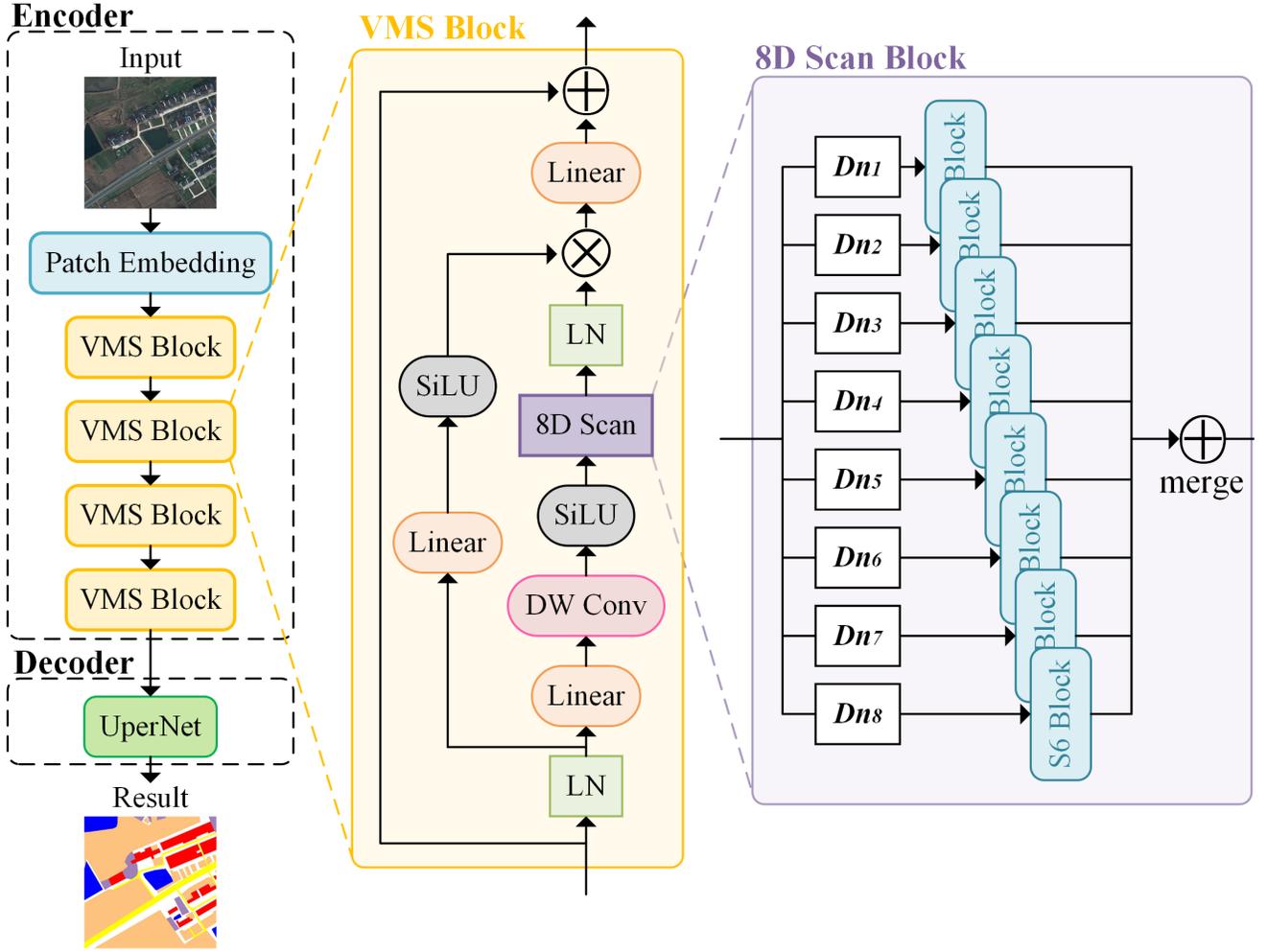

**Fig. 3.** Overall architecture of the experimental framework.

Although similar to Mamba, this architecture exhibits a key difference in the form of image scanning, referred to as the 8-Direction Scan (8D Scan) Block, as shown on the right-hand side of Fig. 3. As the number of considered scanning directions in our experiments ranges from 1 (i.e., unidirectional) to 8 (i.e., a combination of 8 individual scanning directions), we designed 8 potential scanning directions within the 8D Scan Block: *Dn1*, *Dn2*, *Dn3*, ..., *Dn8*. After being separately processed in each of these 8 potential scanning directions, image patches undergo feature extraction through the S6 Block, and subsequently the features from all eight directions are merged. When the number of considered scanning directions is 1, 2, or 4, the scanning directions are repeated 8, 4, and 2 times, respectively, to fill the eight potential scanning directions.

IV. Experiments

*A. Datasets*

To minimize the potential influence of varying characteristics of different datasets on results, we investigated the impact of scanning strategies across three commonly used semantic segmentation datasets in remote sensing, namely ISPRS Vaihingen, ISPRS Potsdam, and LoveDA. In our experiments, the dataset settings are consistent with those widely used in other studies [5, 20], as detailed below.

ISPRS Vaihingen consists of 33 high-resolution remotely sensed images, featuring a spatial resolution of 9 cm and varying image sizes (on average 2494×2064 pixels). These images cover near-infrared, red, and green bands, and are categorized into six classes: impervious surface, building, low vegetation, tree, car, and

clutter. Images labeled with IDs 1, 3, 5, 7, 11, 13, 15, 17, 21, 23, 26, 28, 30, 32, 34, and 37 are used for training, while the remaining 17 images are used for validation.

ISPRS Potsdam has the same categories as ISPRS Vaihingen but features a spatial resolution of 5 cm. This dataset contains 38 images, each with an identical image size of 6000×6000 pixels. It covers four spectrum bands: red, green, blue, and near-infrared, with only the RGB channels being used in our study. Images with the following IDs 2_10, 2_11, 2_12, 3_10, 3_11, 3_12, 4_10, 4_11, 4_12, 5_10, 5_11, 5_12, 6_07, 6_08, 6_09, 6_10, 6_11, 6_12, 7_07, 7_08, 7_09, 7_10, 7_11, and 7_12 are used for training. The remaining 14 images are used for validation. Consistent with ISPRS Vaihingen, the clutter category is excluded from the result evaluation.

LoveDA [21] contains 1669 validation images, 1796 test images, and 2522 training images. All images have a size of 1024×1024 pixels, with a spatial resolution of 30 cm, covering 7 categories: background, building, road, water, barren, forest, and agricultural. The validation set is used for performance evaluation in our study.

TABLE I
TRAINING SETTING FOR SEMANTIC SEGMENTATION NETWORKS

|  | LoveDA | Vaihingen | Potsdam |
| --- | --- | --- | --- |
| Resize | 2048×512 | 512×512 | 1024×1024 |
| Crop size |  | 512×512 |  |
| Total learning iterations |  | 30000 |  |
| Batch size |  | 8 |  |
| Optimizer |  | AdamW |  |
| Weight decay |  | 0.01 |  |
| Schedule |  | PolyLR |  |
| Warmup |  | 1500 iterations |  |
| Learning rate |  | 0.0003 |  |
| Loss function |  | Cross entropy |  |

*B. Training Settings*

In this study, we utilized widely adopted training settings to ensure an effective comparison, as detailed in TABLE I. Due to the limited number of training sets available, data augmentation is employed to prevent overfitting [35]. Random resize, random crop, random flip, and photometric distortion are consistently applied for data augmentation in our experiments. Experiments are performed using two RTX 4090D GPUs.

TABLE II
EFFECTS OF PATCHING SIZE AND STRIDE ON SEGMENTATION ACCURACY IN TERMS OF mIoU.

| Patch size | Stride | Flops | #Param. | Vaihingen | Potsdam | LoveDA |
|---|---|---|---|---|---|---|
| 4×4 | 4 | 252.00G | 79.20M | 74.73 | 82.88 | 46.25 |
| 8×8 | 4 | 246.00G | 79.22M | 72.87 | 81.86 | 45.21 |
| 8×8 | 8 | 63.16G | 79.22M | 69.08 | 79.78 | 43.15 |
| 16×16 | 4 | 240.00G | 79.27M | 70.86 | 79.97 | 43.39 |
| 16×16 | 8 | 60.26G | 79.27M | 66.30 | 77.43 | 38.52 |
| 16×16 | 16 | 15.86G | 79.27M | 60.57 | 74.23 | 34.40 |
| 32×32 | 8 | 57.78G | 79.49M | 61.27 | 71.99 | 34.11 |
| 32×32 | 16 | 14.55G | 79.49M | 55.29 | 69.80 | 31.60 |
| 32×32 | 32 | 4.04G | 79.49M | 47.54 | 63.11 | 31.17 |

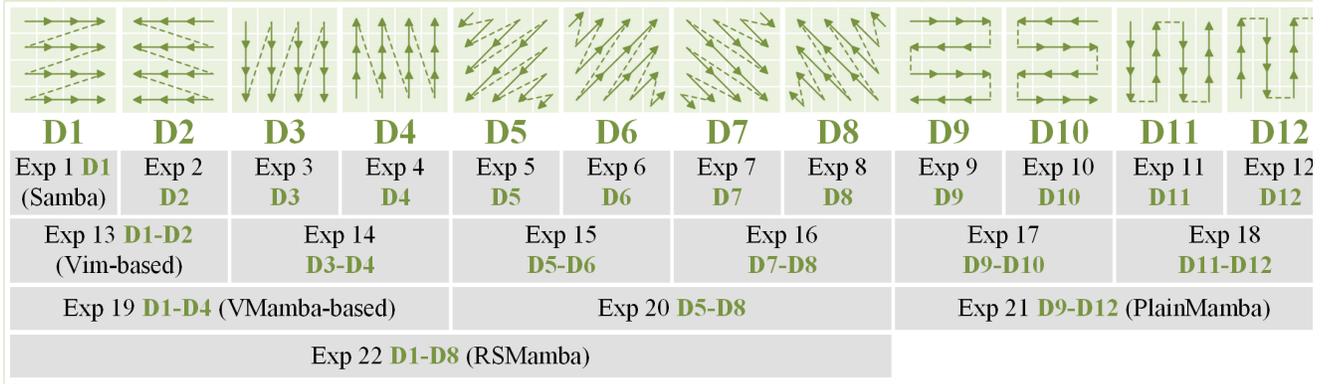

Fig. 4. Experiment design for scanning directions and their combinations.

### C. Patching Methods

The patch size employed for image cropping and the stride used during scanning can also affect experimental results. Currently, most Mamba-based visual tasks adopt a patch size of 4×4 and a stride of 4. To explore the most suitable patch size and stride for subsequent experiments, this study carries out ablation experiments of various patching methods across three considered datasets. To quantify the impact of patching methods on segmentation, the Mean Intersection over Union (mIoU) is used as the metric to demonstrate the segmentation accuracy in the ablation tests. For consistent comparison, the D1 scanning direction is used in the ablation experiments. The stride also plays a crucial role as it affects the sequence's length and computational load. We use floating point operations (Flops) to quantify the computational load, which is calculated using one random-generated 512×512 image as input. Doubling the stride reduces the number of pixels to compute by four times, resulting in an approximate fourfold computational reduction. Considering practicality, the minimum stride in our experiments is set at 4, because smaller strides would demand excessively high computational resources, making training impractical on two 24G GPUs. Patch sizes considered in our experiments include 4×4, 8×8, 16×16, and 32×32, each paired with a stride of the same width as the corresponding patch size for segmentation of a whole image. Additionally, strides smaller than patch sizes are also considered to allow for overlapping scanning of images.

*D. Experiments Design*

Fig. 4 shows 22 scanning strategies tested in our experiments. Experiments (Exp) 1 to 12 consist of individual directional scans, where each of the eight potential scanning directions in the 8D Scan Block is set to that single direction. Exp 13 to 18 represent six sets of bidirectional scanning experiments, where the two directions are repeated four times to fill the eight potential scanning directions in the 8D Scan Block. Exp 19 to 21 consist of three sets of experiments for four-directional scans, in which four scanning directions are repeated twice in the 8D Scan Block. For example, in Exp 19, the eight potential scanning directions are filled with four scanning directions repeated twice. From Dn1 to Dn8, they are D1, D2, D3, D4, D1, D2, D3, and D4. Exp 22 involves combining all eight scanning directions and inputting them to the 8D Scan Block. This arrangement ensures that these experiments' parameters and computational load remain consistent. We use the mIoU metric to assess the overall effectiveness of segmentation, and the IoU score for each class, across the three datasets.

Scanning strategies (as shown in Fig.4) investigated in our experiments include those already adopted in previous studies. Specifically, the scanning strategies in Exp 1, Exp13, Exp19, Exp21 and Exp22 correspond to those used in Samba [20], Vim-based [17], VMamba-based [16], PlainMamba [18], and RSMamba [19], respectively.

TABLE III
EFFECTS OF SCANNING STRATEGIES ON SEGMENTATION ACCURACY FOR ISPRS VAIHINGEN, INCLUDING MIOU AND IOU FOR INDIVIDUAL CLASSES

|  |  | *mIoU* | impervious surface | building | low vegetation | tree | car |
|---|---|---|---|---|---|---|---|
| Exp 1 | Unidirectional | 74.73 | 83.17 | 88.05 | 68.98 | 78.44 | 55.01 |
| Exp 2 |  | 74.83 | 81.82 | 88.94 | 67.54 | 77.80 | 58.05 |
| Exp 3 |  | 74.79 | 82.52 | 88.64 | 67.31 | 77.66 | 57.81 |
| Exp 4 |  | 74.76 | 82.28 | 88.66 | 66.97 | 77.99 | 57.90 |
| Exp 5 |  | 74.91 | 82.73 | 89.38 | 67.56 | 78.02 | 56.84 |
| Exp 6 |  | 74.47 | 81.99 | 88.80 | 66.78 | 77.92 | 56.88 |
| Exp 7 |  | 75.04 | 82.67 | 89.38 | 67.46 | 77.94 | 57.74 |
| Exp 8 |  | 74.65 | 81.99 | 88.79 | 67.00 | 78.23 | 57.23 |
| Exp 9 |  | 74.60 | 82.71 | 89.29 | 67.81 | 77.92 | 55.26 |
| Exp 10 |  | 74.66 | 82.02 | 89.30 | 67.29 | 77.89 | 56.78 |
| Exp 11 |  | 75.02 | 82.77 | 89.20 | 67.52 | 78.02 | 57.61 |
| Exp 12 |  | 75.15 | 82.89 | 89.17 | 67.52 | 77.98 | 58.21 |
| Exp 13 | Bidirectional | 74.34 | 82.19 | 89.25 | 67.05 | 77.79 | 55.43 |
| Exp 14 |  | 74.90 | 82.59 | 89.34 | 67.30 | 78.04 | 57.25 |
| Exp 15 |  | 74.88 | 82.56 | 89.36 | 67.75 | 78.20 | 56.54 |
| Exp 16 |  | 74.78 | 82.58 | 89.06 | 67.35 | 77.83 | 57.07 |
| Exp 17 |  | 74.19 | 82.18 | 88.64 | 66.94 | 77.92 | 55.27 |
| Exp 18 |  | 75.62 | 82.98 | 89.52 | 68.08 | 78.11 | 59.42 |
| Exp 19 | Four-directional | 74.74 | 82.17 | 89.27 | 66.99 | 77.68 | 57.59 |
| Exp 20 |  | 74.91 | 82.39 | 88.93 | 67.67 | 78.10 | 57.45 |
| Exp 21 |  | 74.76 | 82.78 | 89.23 | 67.43 | 77.78 | 56.58 |
| Exp 22 | Eight-directional | 75.24 | 82.83 | 89.46 | 67.30 | 77.77 | 58.86 |

10TABLE IV
EFFECTS OF SCANNING STRATEGIES ON SEGMENTATION ACCURACY FOR ISPRS POTSDAM, INCLUDING mIoU AND IoU FOR INDIVIDUAL CLASSES

|  |  | *mIoU* | impervious surface | building | low vegetation | tree | car |
|---|---|---|---|---|---|---|---|
| Exp 1 |  | 82.88 | 84.49 | 91.08 | 74.82 | 75.69 | 88.31 |
| Exp 2 |  | 83.21 | 84.71 | 91.25 | 75.52 | 75.86 | 88.70 |
| Exp 3 |  | 82.75 | 84.84 | 90.84 | 74.83 | 75.58 | 87.68 |
| Exp 4 |  | 83.07 | 84.90 | 91.39 | 74.98 | 75.89 | 88.20 |
| Exp 5 |  | 83.65 | 85.46 | 91.38 | 76.13 | 76.34 | 88.95 |
| Exp 6 | Unidirectional | 83.39 | 85.22 | 91.46 | 75.44 | 76.31 | 88.51 |
| Exp 7 |  | 83.43 | 85.23 | 91.52 | 75.97 | 76.20 | 88.24 |
| Exp 8 |  | 83.28 | 85.01 | 90.97 | 75.85 | 76.49 | 88.06 |
| Exp 9 |  | 83.04 | 84.73 | 90.74 | 75.46 | 75.92 | 88.36 |
| Exp 10 |  | 83.12 | 84.98 | 90.87 | 75.48 | 75.86 | 88.43 |
| Exp 11 |  | 83.57 | 85.45 | 91.92 | 75.74 | 75.94 | 88.78 |
| Exp 12 |  | 83.32 | 85.46 | 91.47 | 75.18 | 75.98 | 88.52 |
| Exp 13 |  | 82.62 | 84.57 | 90.47 | 74.85 | 75.06 | 88.15 |
| Exp 14 |  | 82.94 | 85.15 | 91.21 | 75.09 | 75.16 | 88.08 |
| Exp 15 | Bidirectional | 83.04 | 84.95 | 91.20 | 75.05 | 75.77 | 88.23 |
| Exp 16 |  | 82.93 | 84.82 | 91.24 | 75.16 | 75.38 | 88.04 |
| Exp 17 |  | 82.68 | 84.22 | 90.81 | 74.90 | 74.73 | 88.74 |
| Exp 18 |  | 82.33 | 84.38 | 91.05 | 74.31 | 74.42 | 87.50 |
| Exp 19 | Four-directional | 82.94 | 84.40 | 90.97 | 75.43 | 75.42 | 88.48 |
| Exp 20 |  | 83.21 | 85.03 | 91.12 | 75.32 | 75.70 | 88.88 |
| Exp 21 |  | 83.08 | 84.93 | 91.27 | 75.65 | 75.07 | 88.49 |
| Exp 22 | Eight-directional | 83.05 | 84.63 | 91.51 | 75.46 | 75.53 | 88.12 |

## V. RESULTS

### A. Impact of Patching Methods

The segmentation accuracy resulted from various patching sizes and strides is presented in TABLE II. A consistent finding was seen from the performance analysis across three datasets: when processing image input sizes of 512×512, a 4×4 patch size with a stride of 4 yields the highest segmentation accuracies for all three datasets. While decreasing the stride sequentially to 4 and maintaining a fixed patch size, Mamba did not exhibit bottlenecks in handling these long sequences, indicating its potential to manage even smaller strides efficiently. Therefore, Mamba shows promise in handling long sequences with smaller strides. When the stride was fixed, reducing patch size improved segmentation performance, suggesting that the Mamba architecture was more adept at processing finer image patches. Based on these findings, we consistently used a 4x4 patch size and a stride of 4 in subsequent experiments (i.e., Exp 1 - Exp 22).

### B. Impact of Scanning Strategies

TABLE III, IV and V present the semantic segmentation accuracies for ISPRS Vaihingen, ISPRS Potsdam, and LoveDA, respectively, using the 22 scanning strategies detailed in Fig. 4.

11TABLE V
EFFECTS OF SCANNING STRATEGIES ON SEGMENTATION ACCURACY FOR LOVEDA, INCLUDING mIoU AND IoU FOR INDIVIDUAL CLASSES

|  |  | mIoU | background | building | road | water | barren | forest | agricultural |
|---|---|---|---|---|---|---|---|---|---|
| Exp 1 | Unidirectional | 46.25 | 51.62 | 57.79 | 51.58 | 57.53 | 21.34 | 38.98 | 44.90 |
| Exp 2 |  | 47.46 | 52.75 | 57.41 | 51.44 | 59.96 | 24.84 | 39.47 | 46.37 |
| Exp 3 |  | 47.54 | 51.47 | 57.47 | 51.20 | 61.85 | 23.43 | 40.90 | 46.46 |
| Exp 4 |  | 46.83 | 50.94 | 58.14 | 52.03 | 59.74 | 23.74 | 40.75 | 42.44 |
| Exp 5 |  | 46.61 | 51.75 | 59.14 | 51.60 | 55.10 | 24.92 | 38.59 | 45.20 |
| Exp 6 |  | 47.05 | 52.10 | 59.39 | 50.59 | 54.89 | 26.97 | 39.84 | 45.56 |
| Exp 7 |  | 46.35 | 50.51 | 56.51 | 52.17 | 57.17 | 22.94 | 40.33 | 44.80 |
| Exp 8 |  | 47.42 | 51.61 | 58.55 | 52.03 | 60.01 | 26.30 | 39.43 | 44.04 |
| Exp 9 |  | 47.71 | 52.31 | 59.52 | 52.07 | 58.71 | 25.91 | 40.10 | 45.35 |
| Exp 10 |  | 47.19 | 50.68 | 56.23 | 52.20 | 61.65 | 23.26 | 40.20 | 46.10 |
| Exp 11 |  | 46.80 | 51.10 | 57.82 | 52.59 | 56.61 | 26.30 | 39.43 | 43.74 |
| Exp 12 |  | 46.99 | 50.97 | 57.22 | 53.54 | 58.60 | 23.39 | 40.85 | 44.39 |
| Exp 13 | Bidirectional | 47.90 | 51.11 | 57.49 | 52.17 | 62.89 | 24.90 | 40.43 | 46.30 |
| Exp 14 |  | 46.88 | 52.07 | 55.41 | 51.83 | 58.30 | 26.79 | 38.88 | 44.88 |
| Exp 15 |  | 47.41 | 51.40 | 57.14 | 53.13 | 61.51 | 24.27 | 38.41 | 46.01 |
| Exp 16 |  | 47.29 | 51.24 | 57.45 | 52.80 | 61.31 | 24.36 | 39.49 | 44.39 |
| Exp 17 |  | 47.39 | 51.73 | 56.19 | 51.63 | 61.12 | 26.52 | 39.57 | 44.96 |
| Exp 18 |  | 46.11 | 51.23 | 56.83 | 51.27 | 55.73 | 27.63 | 36.55 | 43.55 |
| Exp 19 | Four-directional | 46.96 | 50.48 | 54.39 | 53.84 | 57.35 | 27.25 | 40.28 | 45.12 |
| Exp 20 |  | 47.69 | 50.97 | 59.11 | 51.90 | 57.10 | 30.34 | 40.63 | 43.80 |
| Exp 21 |  | 45.80 | 50.75 | 48.33 | 53.37 | 57.64 | 25.89 | 39.15 | 45.45 |
| Exp 22 | Eight-directional | 47.59 | 52.31 | 57.60 | 54.34 | 57.89 | 24.75 | 39.14 | 47.13 |

Analyzing the outcomes of these experiments, we observed an interesting phenomenon across all three datasets used in our experiments. The segmentation accuracies resulting from the 22 scanning strategies seemed to be similar. Takin into account small performance differences among different scanning strategies within each dataset, as well as performance variations of a single scanning strategy across all three datasets, there was no apparent indication of any specific scanning strategy outperforming others, regardless of their complexity or involvement of single or multiple scanning directions. Any slight performance fluctuations observed were probably attributable to the randomness of the training process.

## VI. DISCUSSION AND FUTURE WORK

For semantic segmentation of high-resolution remotely sensed images, our study finds that the utilization of particular scanning directions or combinations of different scanning directions, as proposed in existing Mamba-based approaches, does not effectively improve segmentation accuracy. Therefore, in the Vision Mamba framework, using a ViT-like flattening approach (i.e., D1 scanning) remains effective for semantic segmentation of such images. Moreover, employing a unidirectional scanning strategy such as D1 also reduces computational demands, allowing for deeper network stacking within limited computational resources.

Exploring the generalization of language models for vision tasks is of significant importance for the development of deep learning, as evidenced by the success of ViT [4]. Recent advancements in recurrence-based models [11, 36] suggest the ongoing exploration of effective ways to integrate them into visual tasks. Current efforts focus on designing strategies for scanning image patches to enhance the model's understanding of image sequences. However, our investigation into semantic segmentation of remotely sensed images revealed that Mamba-based models are not sensitive to different scanning strategies. Therefore, it is arguable that directing efforts towards exploring more effective ways, other than different scanning strategies, to enhance the Mamba model's understanding of remotely sensed images.

Our work does not intend to discredit the extensive efforts to improve scanning strategies in Vision Mamba but to demonstrate that these improvements have limited effects on semantic segmentation of remotely sensed images. This phenomenon is explainable: remotely sensed images differ from conventional images in terms of features. On one hand, when comparing with conventional images such as pictures of people, the difference between patches representing the same semantic in remotely sensed images is minimal when converted into sequences. On the other hand, their causal linkage of the sequences is weaker than conventional images.

However, the effectiveness of varying scanning strategies in other types of datasets with more apparent causal relationships in sequences, such as COCO_Stuff [37] and Cityscapes [38], remains to be verified, which presents an interesting area for future work.

During our experimenting with different patching methods, we discovered an interesting phenomenon: reducing the stride improved segmentation accuracy, albeit at the expense of increased computational demands. This suggests that Mamba may perform better with strides smaller than the smallest stride (i.e., 4) used in our experiments for processing images of 512×512 pixels. However, the exponential increase in sequence length with smaller strides has precluded our experimentation with those smaller strides due to our computational resources available. Investigating more efficient computational methods to accommodate denser scanning is a meaningful direction for future study.

## VI. Conclusion

This study quantitively investigated the impact of 22 scanning strategies in the Mamba-based approach for semantic segmentation of high-resolution remotely sensed images, across the ISPRS Vaihingen, ISPRS Potsdam and LoveDA datasets. The experimental outcomes demonstrated that there was no discernible enhancement in segmentation accuracy resulting from the various scanning strategies, whether unidirectional scanning directions or their combinations. Therefore, for remotely sensed images, a simple flattening method was deemed sufficient in Mamba-based approaches. However, it was suggested that the effectiveness of multi-directional scanning methods for conventional images still required validation.

Our study also found that reducing the stride would enhance Mamba's performance in semantic segmentation, but at the cost of heightened computational resources. Therefore, it is valuable to develop more efficient computational methods to support denser scanning.